\pdfoutput=1

\documentclass[11pt]{article}

\usepackage{acl}

\usepackage{times}
\usepackage{latexsym}

\usepackage[T1]{fontenc}
\usepackage{graphicx}

\usepackage{url}
\usepackage{subfig}
\usepackage[utf8]{inputenc}

\usepackage{microtype}

\title{When to Laugh and How Hard?\\A Multimodal Approach to Detecting Humor and its Intensity}

\author{Khalid Alnajjar\textsuperscript{1,2},~Mika Hämäläinen\textsuperscript{1,2}\\
 \bf{Jörg Tiedemann\textsuperscript{1}, Jorma Laaksonen\textsuperscript{3} \and Mikko Kurimo\textsuperscript{3}} \\
        \textsuperscript{1}University of Helsinki, Finland \\
        \textsuperscript{2}Rootroo Oy, Finland \\
  \textsuperscript{3}Aalto University, Finland \\
  \texttt{firstname.lastname@\{helsinki.fi\textsuperscript{1},aalto.fi\textsuperscript{3}\}} \\}

\begin{document}
\maketitle
\begin{abstract}
Prerecorded laughter accompanying dialog in comedy TV shows encourages the audience to laugh by clearly marking humorous moments in the show. We present an approach for automatically detecting humor in the \textit{Friends} TV show using multimodal data. Our model is capable of recognizing whether an utterance is humorous or not and assess the intensity of it. We use the prerecorded laughter in the show as annotation as it marks humor and the length of the audience's laughter tells us how funny a given joke is. We evaluate the model on episodes the model has not been exposed to during the training phase. Our results show that the model is capable of correctly detecting whether an utterance is humorous 78\% of the time and how long the audience's laughter reaction should last with a mean absolute error of 600 milliseconds.
\end{abstract}

\section{Introduction}

Humor is a topic that has piqued interest of the computational creativity research community over the years. There are numerous systems that can generate humor using a variety of different methods~\cite{weller-etal-2020-humor,tylercomputational,3e8cb38e86694ee3adb6108734e5fff4}. But just as important as it is to research generation from the computational creativity perspective, it is to study automatic assessment of humor.

The role of humor is an important one for us humans as it has it's own social function~\cite{ziv2010social}. It helps us talk about difficult topics~\cite{vivona2013investigating,monahan2015use} and relieves tension~\cite{shurcliff1968judged}. Laughter has a role in building relationships~\cite{mccabe2017laughter,kurtz2017sharing} and it has a positive effect on brain chemistry~\cite{gonot2018laughter}. Humor and laughter are therefore an integral part of who we are as a species. 

Humor is a phenomenon that requires surprise and coherence (see~\citet{d76230f1c2ad4e9f87cd5f3840ae2742}); or incongruity and its resolution in other terms~\cite{raskin1984semantic,attardo1991script}. However, what is surprising or incongruous, depends heavily on the context where humour is presented. Quite indeed, something intended as a joke can be seen as a severe insult just by a change in context. Therefore, we believe that a multimodal approach needs to be researched when when assessing humor automatically; something that thus far has been researched by focusing on the textual modality alone.

Annotated multimodal datasets are scarce, but an access to such a dataset is crucial for any computational attempt on humor detection and assessment. For this reason, we embrace a clever approach: we use episodes of the beloved American sitcom \textit{Friends} as our data source. The TV show has prerecorded audience laughter which provides us with an ultimate source for annotations. Every time the audience laughs, we know that there was something humorous immediately before the laughter. A lack of laughter indicates no humor. In addition to this, the audience can laugh for a short or a long time, which allows us to gather data on how funny a given joke was.

We propose a pipeline consisting of two neural models. One of them detects whether a sentence is humorous or not and the other rates how funny the sentence is in case it was deemed humorous. We evaluate our models on episodes of \textit{Friends} that were not used during training or validation. Our work is a first step towards multimodal humor detection and assessment. We have also established several important data processing practices that make it possible for future research to automatically gather annotated multimodal data in a similar fashion as we did in this paper.

The main contributions of our paper are as follows:
\begin{itemize}
 \item Methodology for automatically annotating a multimodal humor corpus based on laughter cues.
 \item A multimodal humor detection model that does not rely on an explicit split in a setup and a punchline.
 \item A multimodal humor assessment model that can predict how funny a given joke is.
\end{itemize}

\section{Related work}

There is a an extensive body of literature that focuses on humor generation~\cite{dybala2010multiagent,08aa62f6e667427db7a8d0c4e11e21b3,mishra-etal-2019-modular,yamane2021humor} and also a growing body of work that deals with multimodality in natural language understanding~\cite{soldner-etal-2019-box,wang-etal-2020-docstruct,rodriguez-bribiesca-etal-2021-multimodal}. However, in this section, we focus on some of the recent papers that deal with humor detection and analysis.

In a recent work~\cite{xie-etal-2021-uncertainty}, the authors study humor detection in a context where there is a setup and a punchline. They use a GPT-2 model~\cite{radford2019language} to assess uncertainty and surprise to determine if a setup-punchline pair is a joke or not. Similar setup and punchline based approaches for humor detection have been widely studied in the past using different computational methodologies~\cite{cattle-ma-2016-effects,cattle-ma-2018-recognizing,wang-etal-2020-unified}. However, such methods are very different from our approach as we do not expect our model to receive a setup and a punchline that are explicitly marked in the data.

Sentiment analysis has been used in humor detection~\cite{liu-etal-2018-modeling}. As many of the existing approaches, their approach also operates on a setup and a punchline. The authors found that sentiment conflict and transition between the setup and the punchline are useful in humor detection. The authors use an existing discourse parser~\cite{feng2012text} combined with TexBlob\footnote{https://github.com/sloria/textblob} sentiment analysis and heuristic rules to detect humor.

Apart from humor detection, there is a line of work on assessing the humor value of a joke~\cite{weller-seppi-2019-humor}. The authors propose a model that does not detect humor, as it expects jokes as its input, but instead, the model rates how humorous a given input joke is. Their model also expects a setup and a punchline division in the data. The authors train a transformer~\cite{vaswani2017attention} based model for the task.

Humorous headlines have been automatically ranked based on how funny they are~\cite{dick-etal-2020-humoraac}. The authors use ridge regression and an LSTM (Long short-term memory) model with manually engineered features such as whether Donald Trump and his hair have been mentioned and the length of the headline. The authors conclude that a language model is simply not enough for assessing humor, but a wider context is needed to help the model understand humor.

As we can see, most of the previous approaches on humor detection and assessment do either or. The models can either tell whether something is funny or not, or rate how funny a given joke is. In addition, many models seem to expect a clear division into a setup and a punchline, which makes it impossible to use them to detect humor in free formed speech or text. In addition, there are many types of humor, for instance sarcastic one~\cite{e5f27c84c8a14c44bbbbe789d9868b1f}, that does not require an explicitly uttered setup in natural language, but rather the setup of the joke can be deduced from the context itself. Our approach tries to tackle these shortcomings in the current state of humor detection.

\section{Humor theories}

Humor is an integral part of our social lives as humans and because of that, it has provoked the interest of many scholars in the past.  Some of the early theories of humor \cite{hobbes1904leviathan} saw it as a question of superiority, where a superior person laughs at the misery of those inferior to them. While this explanation might be valid in the context of schadenfreude, more modern takes on humor theory reject it as it cannot explain humor as a whole.

For \citet{koestler1964act}, humor is a part of creativity together with other two components: discovery and art. What is seen as characteristic to humor, in his view, in comparison to the other two constituents of creativity, is that its emotional mood is of an aggressive nature. Humor comes from bisociation which is a collision of two frames of reference happening in a comic way.

\citet{raskin1984semantic} presents a theory that is quite similar to the previously described one in the sense that in order for a linguistic expression to be humorous, it has to be compatible with two different scripts. The different scripts have to somehow oppose one another, for example in the sense that one script is a real situation and the other is hypothetical. 

\citet{attardo1991script} sees humor to be consisting of six hierarchical resources of knowledge: language, narrative strategy, target, situation, logical mechanism and script opposition. Similarly to the previous theories, the incongruity of two possible interpretations is considered to be an important aspect of humor. An interesting notion that sets this theory apart from others is that of target. According to the authors, it is not uncommon for a joke to have a target, such as an important political person or an ethnic group, to be made fun of.

Two requirements have been suggested in the past as components of humor in jokes: surprise and coherence (see \citet{brownell1983surprise}). A joke will then consist of a surprising element that will need to be coherent in the context of the joke. This is similar to having two incongruous scripts being simultaneously possible.

\citet{veale2004incongruity} discusses that the theories of \citet{raskin1984semantic} and \citet{attardo1991script} entail that people are forced into resolution of humor. He argues that humor should not be seen as resolution of incompatible scripts, but rather as a collaboration, where the listener willingly accepts the humorous interpretation of the joke. Moreover, he argues that while incongruity contributes to humor, it does not alone constitute it.

\section{Data construction}

We focus on the sitcom TV show \textit{Friends}. The show is one of the most popular American sitcoms ever produced and it aired from 1994 to 2004. Our data consists of the entire show, i.e. 10 seasons and 236 episodes each of a duration around 20 minutes. All episodes had English well-aligned subtitles that correspond to what is said in the audio track of each episode. We have randomly sampled an episode from each season to assess the quality of the subtitles, and found no major errors or clear delays.

While there are some multimodal annotated data for sarcasm detection~\cite{castro-etal-2019-towards,alnajjar-hamalainen-2021-que}, multimodal annotated data of humor in more general terms is very scarce. While several textual humor datasets exist~\cite{hossain2019president,meaney-etal-2021-semeval}, they are not suitable for our need as we are interested in multimodality. This is mainly due to the great subjectivity of what humans deem to be funny, and to the high amount of work and funds needed to manually annotate a dataset. To overcome this obstacle, we embrace an automatic approach for annotating humor in the TV show by recognizing laughter in the audience as described in the following subsection.

\subsection{Data annotation}

After the first few seasons, \textit{Friends} was shot entirely in front of a live audience and a great deal of the laughter in the aired version was original, which would even cause the cast to panic when no laughter is heard while it was expected~\cite{friends_reunion}. This makes this show, and other sitcom shows that are shot live a mine of humor annotations that is calling for extraction, given that the laughter is an indication of truly landing jokes rather than being something cued in or added later in the post-editing phase.

Our approach for annotating the show relies on the model proposed by~\citet{gillick21_interspeech} for automatically detecting laughter. The model is designed to be robust enough to work on real data and be capable of detecting laughter "in the wild". On a lower level, the model's implementation relies on ResNet~\cite{he2016deep}. The model allow us to indicate the minimum length of the laughter and a cutoff threshold of how confident the model is. We set the minimum laughter length to 0.2 seconds and the threshold to 80\% based on our empirical experiments. A shorter length resulted in numerous non-continuous short segments of laughter while a longer length limited the results to a few segments per episode. The case was similar for the confidence threshold. A shorter duration for laughter also lead to many non-laughter noises such as yelling to be detected falsely as laughter.

We ran the model on all the episodes of every season of the entire show and obtained 7422 laughter segments. To construct a dataset for training a supervised neural model for identifying humor, positive and negative samples are required. A crucial aspect of what makes something funny is the context it is present in~\cite{Tsakona+2020+7+18}, just like how the utterance ``You guys just keep getting cooler and cooler'' can indicate the opposite of what is expressed and be sarcastic based on the tone of the speaker and what the ``guys'' have done or said\footnote{The utterance was said by Chandler sarcastically in the \textit{The One Where the Stripper Cries} episode.}. Furthermore, the context prompting the humorous interpretation can vary in nature, especially that we are dealing with multi-modality. In our case, it could vary, for instance, based on its length (i.e., how distant is the required knowledge for the act to be deemed funny? A scene, episodes or seasons?), type (i.e., what aspects in the context contribute to the humor? does the humorousness arise from what is said, how it was said or what is done, or a combination?), and familiarity with the characters' personalities and common knowledge of the topic of the discourse. 

\begin{figure}[ht]

\centering
\includegraphics[width=7.5cm]{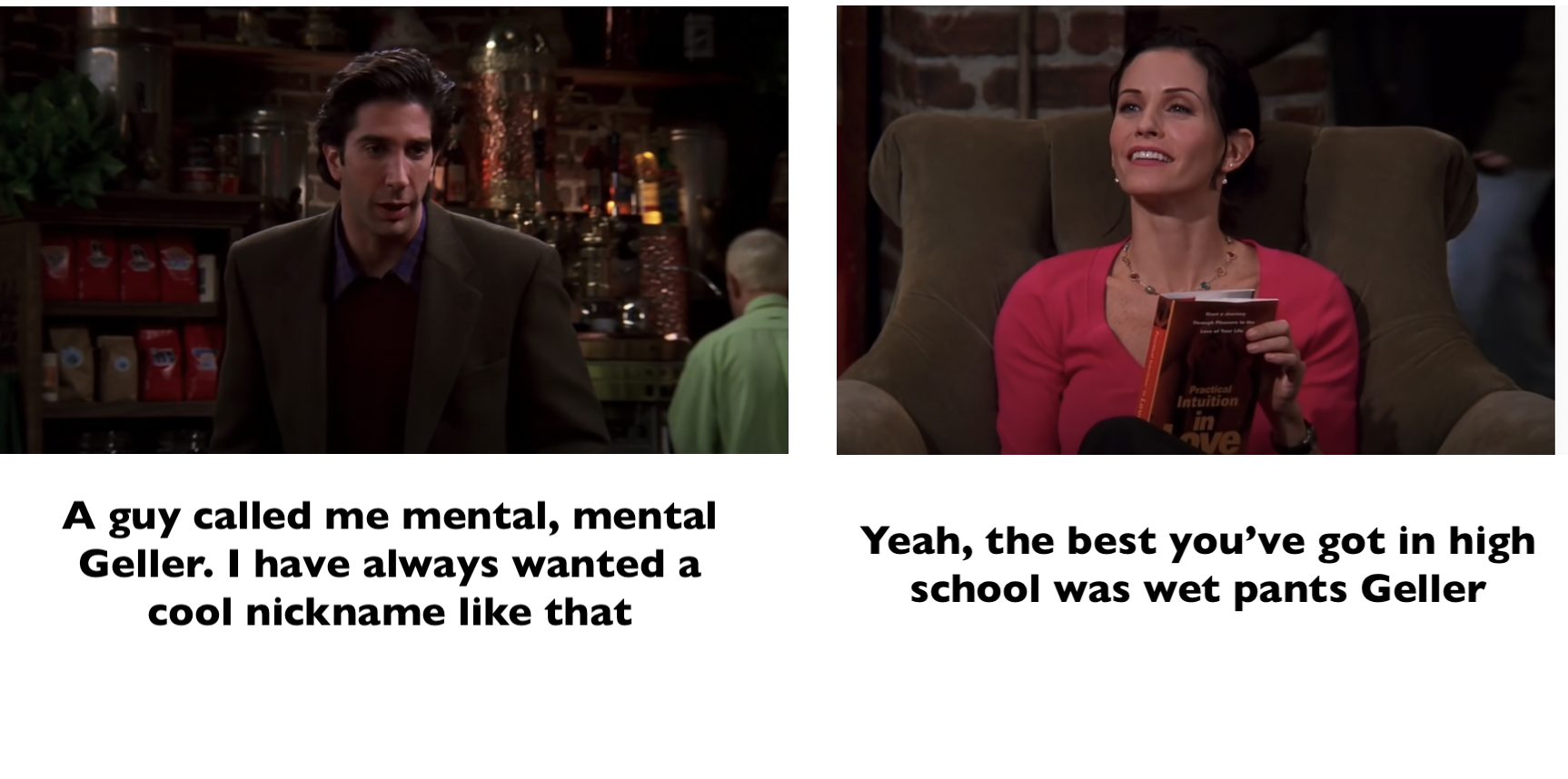}
\caption{An example of a humorous segment that preceded laughter}
\label{friends_joke}
\end{figure}

As it is unfeasible for an unsupervised automated approach to achieve an understanding of the world and, therefore, explain the humor in a given scene and link all contexts in the entire show contributing to it, we resort to defining the context as a fixed duration for the sake of simplicity. For the positive humorous samples, we consider the last 10 seconds of all the laughter segments detected by the model to be the context of the joke. From our experiments, 10 seconds seemed to include some context (e.g., two characters saying two sentences) and not collide with previous laughter segments. An example of such a humorous segment can be seen in Figure~\ref{friends_joke}.

To build a set of negative samples, for each positive sample segment, we consider what is prior to it until the laughter segment preceding it to be a non-humorous segment.  If possible, we split the segment into 10-second clips and randomly pick 3 clips that has some context. The presence of the context is determined by inspecting the subtitles of the clip. In the case where it was empty, the clip was discarded. This is important to remove segments that have no verbal communication at all such as camera spanning across a scenery in the beginning of a scene. Sometimes, laughter segments are very close to each other that no non-humorous segment exists before a humorous segment.

\begin{figure}[ht]

\centering
\includegraphics[width=5cm]{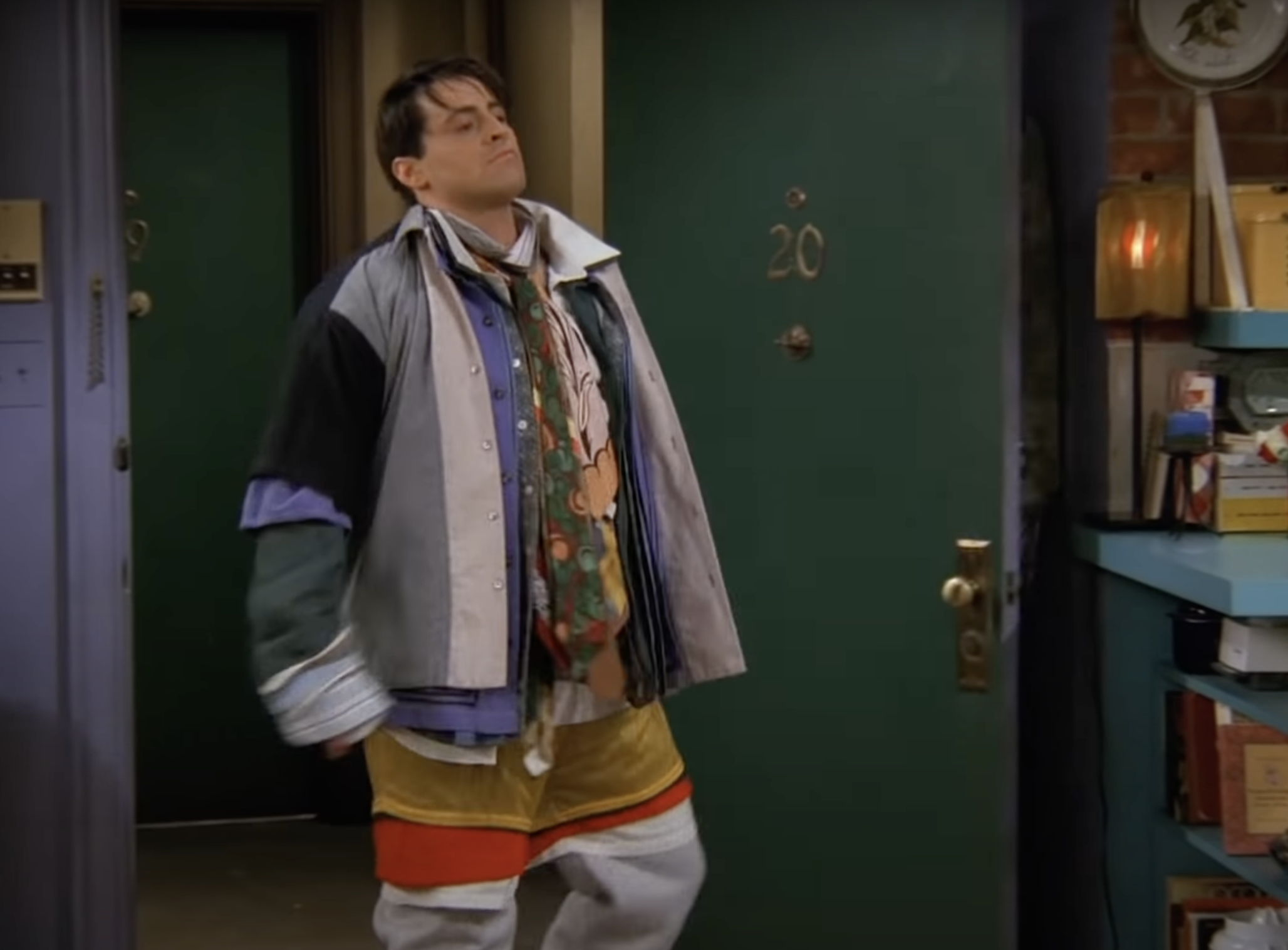}
\caption{Joey entering the room wearing many layers of clothes provoked audience laughter non-verbally}
\label{friends_joey}
\end{figure}

Sometimes humor was expressed non-verbally in the TV show as seen in Figure~\ref{friends_joey}. We filtered out the cases where the audience laughed and there was no subtitle text before the laughter in the humor segment. We do this because we are focusing on multimodality (text and audio) in humor detection. Such non-verbal humour would require the video to be considered as well.

By the end of the annotation process, our dataset consisted of a total of 16710 clips where 7422 of them are humorous and the remaining 9288 are non-humorous clips\footnote{Due copyright we are unable to publicly release the dataset.}. Furthermore, we indicated in the dataset the length of the laughter segment for all humorous clips. This metadata will be used to predict the intensity of the humor.

\subsection{Preprocessing}

In our current work, we focus on text and audio in detecting humor, and leave including visual content for future work. This is due to the fact that processing video requires a research on its own right. Video is such an information rich resource with so many potential things one could extract that may or may not be relevant for humor such as facial expressions, body poses, object recognition, action detection and so on. For our text and audio modalities, two types of preprocessing are applied: 1) cleaning and reformatting the subtitles\footnote{This step improved the accuracy of our models by 3\%.} and 2) resampling the frequency of the audio clips.

Dashes, \textit{-}, at the beginning of subtitle lines were commonly used to imply that the speaker has not changed. We treat such dashes as noise and prune them out. Furthermore, new lines are usually added in subtitles to facilitate reading them and/or to separate talks by different speakers. 
We substitute new lines with spaces in order to convert the subtitle into complete sentences. Italic tags, ``$<$i$>$'' and ``$<$/i$>$'', were stripped out. When inspecting the subtitles, we noticed that a frequent typo of having a capital I instead of an l existed. We addressed this issue by replacing all \textit{I}s with \textit{l}s if they were happened to be in the middle of the word. In case a clip contained multiple subtitle segments across the time-span of the clip, they were all joined together. 

In terms of audio processing, we only apply frequency resampling to adjust the frequency to 16kHz. This step is performed to ensure consistency when feeding the data to a neural model.

\section{Assessing humor}

We present a pipeline of two neural models for assessing humor. The goal of the first model is to identify whether an utterance at the end of the segment is humorous or not given the rest of the segment as context, whereas the second model predicts the intensity of the humor once it has been detected.

\subsection{Humor detection}

Here, we describe two different models for detecting humor. The first model relies solely on the textual data, and the second is a multimodal model that accepts both textual data and audio signals as input.The task for these models is a binary classification downstream task which is to determine whether the input contains humor or not. We experiment with the two models to gain a better understanding on what the effect of the audio is for humor detection.

We group our dataset by episodes and, then, randomly decide which episodes will be used for training, validation and testing. This division is conducted to prevent the model from getting exposed to shared contexts during the training and testing phases, which would introduce undesired bias. When test data is sampled from completely different episodes than what the training and validation contained, we can ensure that the model learns to detect humor in completely novel contexts rather than detecting merely episode specific recurring jokes. The test dataset is constituted of 25 full episodes, which are 1x09, 2x06,
2x22, 3x13, 3x20, 4x09,
5x04, 5x07, 6x09, 6x11,
6x15, 6x16, 7x09, 7x16,
7x19, 7x20, 7x22, 8x03,
8x14, 8x21, 9x05, 9x11,
9x21, 10x17 and
10x18. In total, the training, validation and testing splits contained 13506, 1477 and 1708 samples, respectively. Both of the models used the same splits to ensure comparable results.

\subsubsection{Text only model}
We build our text only model by applying transfer-learning and fine-tuning a BERT model~\cite{devlin-etal-2019-bert} using the transformers Python library~\cite{wolf-etal-2020-transformers}. The pretrained model we used is the uncased English BERT model\footnote{\url{https://huggingface.co/bert-base-uncased}}. 
For a given input, it is first tokenized using BERT tokenization. If the input contained subtitles from different scenes, they we combined together and separated using the special token ``[SEP]''. 

The architecture of the neural network is composed of the BERT model, a BERT pooler layer, a dropout layer~\cite{JMLR:v15:srivastava14a} and a fully connected dense layer that has two outputs. Once the input has passed through the BERT model, the pooler layer returns the last layer hidden-state of the first token of the input sequence. Dropout is applied on the pooler output with a probability of 20\% to reduce overfitting. The linear layer is introduced so that the network would learn a way of interpreting the features produced by the past layers and assign a probability score for each of the two labels. In total, the model has 109 million trainable parameters. We utilize Adam optimizer~\cite{kingma2014adam} with a learning rate of $1\mathrm{e}{-4}$, along with the cross-entropy loss function to optimize the neural network. The fine-tuning process was run for 3 complete epochs.

\subsubsection{Text and audio model}

The multimodal model we propose utilizes the textual and audio input by combining BERT~\cite{devlin-etal-2019-bert} with Facebook's HuBERT\footnote{\url{https://huggingface.co/facebook/hubert-xlarge-ls960-ft}}~\cite{9585401} neural models. We use the same uncased English BERT model in our multimodal model as we did in our text only model to examine the effect of incorporating audio features for detecting humor. The choice of HuBERT, in contrast to the popular XLSR-Wav2Vec2~\cite{conneau21_interspeech}, is due to its superior or, in worse case scenario, neck-to-neck performance.

Our multimodal model architecture is similar to a siamese neural network architecture in the sense that the output of two models are considered collectively. In our model, one side of the network is dedicated to text and the other to audio. We ensure that both sides produce an equal size of features by 1) setting a fixed input length to BERT where padding and truncating is applied where necessary and 2) having two average pooling layers following the output of each side. For the textual output, a global average pooling is applied, whereas an adaptive average pooling is applied to the audio output. Afterwards, the pooled output is concatenated and followed by a dropout layer with a probability of 20\%. Lastly, a fully connected dense layer is employed as the classification layer. The network has 424 million trainable parameters. We use the same hyperparameters for optimizing the multimodal model as the text only model; in other words, we fine-tuned it for 3 full epochs with a learning rate of $1\mathrm{e}{-4}$.

\subsection{Predicting laughter intensity}

We define the intensity of the laughter based on its duration. Thus, a strong laughter for multiple seconds indicates a great joke. This is intuitive because the funnier the humor gets, the longer the audience laughs. As the duration of a laughter is a continuous value, we treat the task a regression problem and adopt an artificial intelligence neural network for addressing it.

Our dataset for this part is only the humorous clips, a total of 7422 clips. Table~\ref{tab:length-dist} shows the number of humorous clips grouped by various durations. 
We cap all the durations to 3 seconds given that the majority of laughter segments are within this limit. The data is then split for training, validation and testing with 80\%, 10\%, and 10\% ratios, respectively.

\begin{table}[]
\centering
\resizebox{\columnwidth}{!}{
\begin{tabular}{|l|c|c|c|c|c|c|c|c|}
\hline
\textbf{Length}   & \textbf{0.2-0.5}  &  \textbf{0.5-1.5}  & \textbf{1.5-2.5}  & \textbf{2.5-3.5}  & \textbf{3.5-4.5}  & \textbf{4.5-5.5}  &  \textbf{5.5-15.5}  \\ \hline
\textbf{N}   & 459  & 2895  & 2328  & 948  & 374  & 184  & 234 \\ \hline
\end{tabular}
}
\caption{The duration of laughter (in seconds) for different ranges in our dataset}
\label{tab:length-dist}
\end{table}

In our laughter intensity prediction model, we only train a multimodal model given that \textit{what} is said in the joke and \textit{how} it is said or performed have great influence on the reaction from the audience. Textual features are extracted using BERT like the aforementioned models. For audio features, we extract them from the entire humorous clips using
Google's VGGish model\footnote{\url{https://github.com/harritaylor/torchvggish}}~\cite{vggish}. VGGish is a Convolutional Neural Network inspired by the VGG network~\cite{Simonyan15} that is trained for image classification. However, in VGGish, the input image is the log mel spectrogram of frames derived from the audio. The network has achieved the state-of-the-art results for audio classification given its ability to capture acoustic features, tones, volume and so on. Unlike the earlier models, the pretrained models, i.e. BERT and VGGish, are frozen and not fine-tuned during the training step.

The model we present here is a sequence of layers where the first one is a 2D average pooling layer that converts the extracted features by BERT and VGGish models into a fixed-size set of features by averaging neighbouring features until the desired size is reached. We set the size here to 128 as this is the size of features that the VGGish model returns per frame. As a result, the output of this layer is a vector of 256 features. The layer is then followed by a dense fully-connected linear layer that takes in the averaged features and learns a new representation of 64 features. ReLU activation \cite{agarap2018deep} and a dropout with a probability of 10\% are then subsequently applied to the 64 features. The network architecture ends with a fully-connected dense layer that returns one output representing the intensity score.

As the problem here is regression, we make use of the mean squared error (squared L2 norm) as the loss function. This model has 16.5 thousand trainable parameters and we optimize them for 100 epochs using Adam optimizer, however we use use early stopping to stop the training of the model before 100 epoch in the event of the model converging early.

\section{Results and evaluation}

We run both, the text only and multimodally trained models, for humor detection on the test split. Their performance is assessed using precision, recall, F1 and accuracy scores, which are given in Table~\ref{tab:models-results}. Both of our models outperform the baselines of choosing a label at random or the most frequent label, their accuracies were 51\% and 56\%, in the order given.

\begin{table}[]
\centering
\begin{tabular}{|lcccc|}
\hline
\multicolumn{1}{|l|}{}         & \multicolumn{1}{c|}{\textbf{Precision}} & \multicolumn{1}{c|}{\textbf{Recall}} & \multicolumn{1}{c|}{\textbf{F1}}   & \textbf{N}   \\ \hline
\multicolumn{5}{|c|}{\textbf{Text only model}}                                                                                           \\ \hline
\multicolumn{1}{|l|}{Funny}    & \multicolumn{1}{c|}{0.58}      & \multicolumn{1}{c|}{0.52}   & \multicolumn{1}{c|}{0.55} & 758 \\ \hline
\multicolumn{1}{|l|}{Not funny} & \multicolumn{1}{c|}{0.65} & \multicolumn{1}{l|}{0.70} & \multicolumn{1}{l|}{0.68} & \multicolumn{1}{l|}{950} \\ \hline
\multicolumn{1}{|l|}{Accuracy} & \multicolumn{4}{c|}{62\%}                                                                       \\ \hline
\multicolumn{5}{|c|}{\textbf{Text + Audio model}}                                                                                        \\ \hline
\multicolumn{1}{|l|}{Funny}    & \multicolumn{1}{c|}{0.69}      & \multicolumn{1}{c|}{0.90}   & \multicolumn{1}{c|}{0.78} & 758 \\ \hline
\multicolumn{1}{|l|}{Not funny} & \multicolumn{1}{c|}{0.90} & \multicolumn{1}{c|}{0.68} & \multicolumn{1}{c|}{0.78} & 950                      \\ \hline
\multicolumn{1}{|l|}{Accuracy} & \multicolumn{4}{c|}{78\%}                                                                       \\ \hline
\end{tabular}
\caption{Accuracy, Precision, recall, and F1 scores of the two models for detecting humor}
\label{tab:models-results}
\end{table}

The results indicate that the multimodal model clearly outperformed the text only model, by a 16\% increase in detection accuracy. This suggests that audio cues were helpful in recognizing humor. For instance, sarcasm and irony are sometimes marked with clear intonations and tones. Both of these phenomena are frequently used in sitcoms for humorous effect, which would aid the model in distinguishing when ``Yeah, right'' is meant as a sincere confirmation or as a sarcastic remark for humorous effect.

Recognizing humor is a challenging task, even for humans and it is no surprise that computational models would struggle. As these models are tested on entirely novel contexts (i.e., new discourses that are not covered during the training phase), the performance achieved by them is impressive. 

To test the model for predicting laughter intensity, we compute the mean absolute error between the predicted intensity and the intensity of the laughter in the dataset. The average mean absolute error was only around 600 milliseconds. This means that the model can predict how long the audience will laugh after a given joke rather accurately, given that there is some flexibility in the duration of the laughter as it is not an absolute measurement of the humor of a joke. With these results, we can say that the model has learned to predict the intensity of humor in a joke well, given that for jokes provoking less laughter, the model predicts a short laughter, and for jokes provoking a lot of laughter, the model predicts a long laughter. Even though the model is not quite accurate in knowing the exact duration of the laughter.

\subsection{Error analysis}

When we look at the results of the models, we can see some cases where both of the models failed at predicting the humor accurately. For example, the following dialog provoked a laughter in the audience:

\begin{itemize}
\setlength\itemsep{0em}
  \item \textbf{Phoebe:} It's amazing. My headache is completely gone. What were those pills called?
  \item \textbf{Monica:} Hexadr\textit{i}n.
  \item \textbf{Phoebe:}  I love you Hexadr\textit{i}n.
\end{itemize}

In this example the humor comes from Phoebe's lack of knowledge of how medicines work as she continues by calling the instructions booklet a story. This is an example of humor that requires some world knowledge and also some understanding of Phoebe's care-free character. Just relying on text in this case or even including the audio does not give the model a context wide enough to reach to a correct interpretation.

In some of these cases where neither of the models predicted the humor right, it is evident that with an access to the video, the model could situate the humor better in the context. The following dialog is an example of such humor:

\begin{figure}[ht]

\centering
\includegraphics[width=5cm]{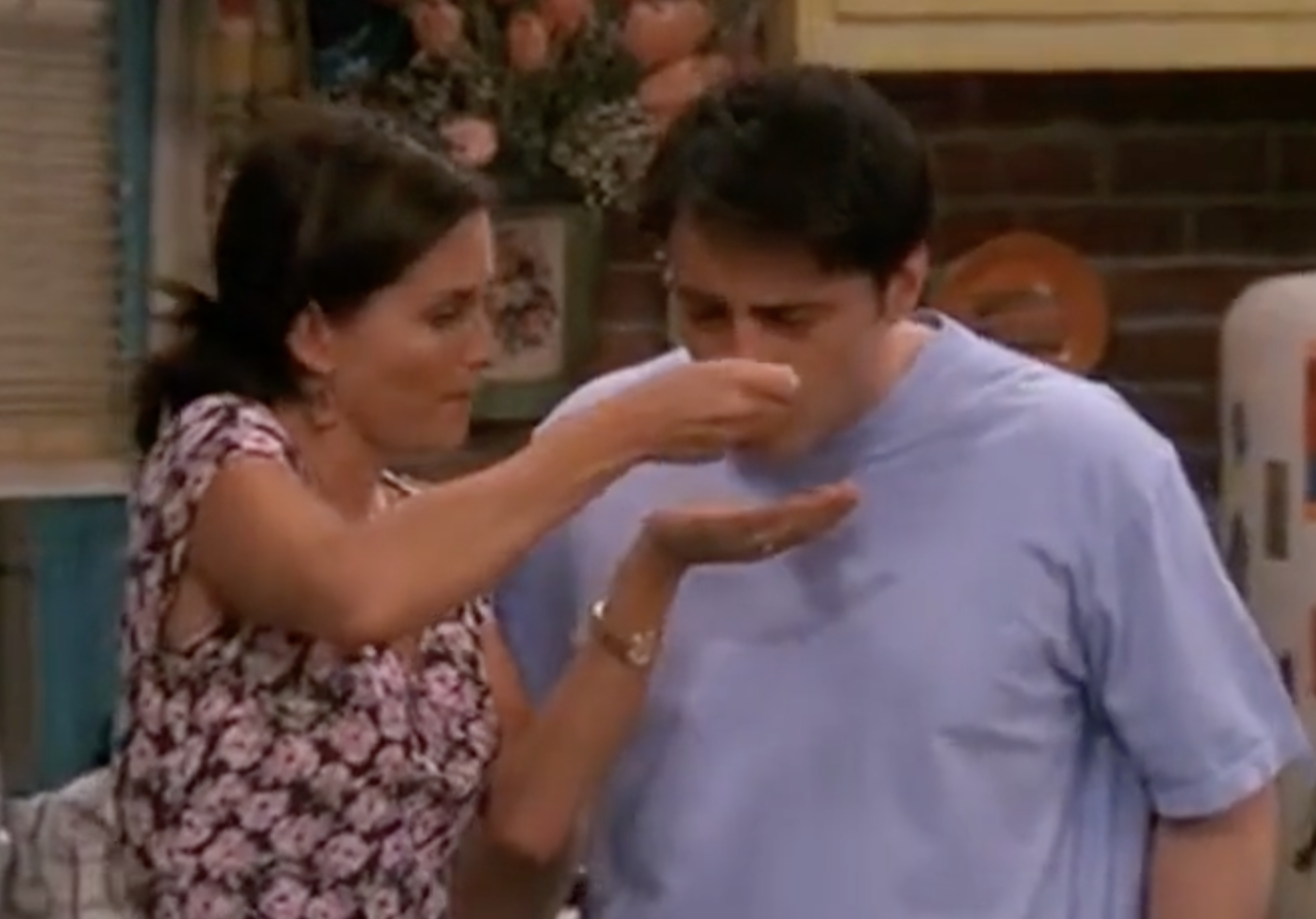}
\caption{Joey inquiring whether Monica had cooked a person after tasting her food.}
\label{friends_joey-c}
\end{figure}

\begin{itemize}
\setlength\itemsep{0em}
  \item \textbf{Monica:} Remember the guy that gave me a bad review? Well... I'm getting my revenge.
  \item \textbf{Joey:} You cooked him?
\end{itemize}

In this dialog, Monica is preparing food and letting Joey taste some of it after her line in the dialog as seen in Figure \ref{friends_joey-c}. A great part of the joke is in the visual action of Joey tasting Monica's food before asking whether she had cooked the guy who had given a bad review. A model that can take the visual modality into account as well could potentially benefit from the humor intensifying action seen on the video.

When we look at the results where the multimodal model predicted humor right and the text only model predicted it falsely as non-humorous, we can see cases where the speaker's use of their voice gave additional context for the humor. An example of such is in the following dialog:

\begin{itemize}
\setlength\itemsep{0em}
  \item \textbf{Katie:} You have selected a lot of nice things. So do you uh, want these things delivered Mr. and Mrs. Geller?
  \item \textbf{Rachel:} Oh
  \item \textbf{Ross:} Oh
  \item \textbf{Rachel:} Oh, no, no. No, no.
\end{itemize}

In this example, the tone of Rachel's voice makes it more evident that Katie was wrong in her assumption that Rachel and Ross were married. The model had learned to capture such a tone of voice as an indication of humor. Another example where the audio is beneficial is the following dialog

\begin{itemize}
\setlength\itemsep{0em}
  \item \textbf{Waiter:} It's just that we do have some large parties waiting.
  \item \textbf{Phoebe:} Oh, one really does have a stick up one's ass, doesn't one?
\end{itemize}

\begin{figure}[ht]

\centering
\includegraphics[width=5cm]{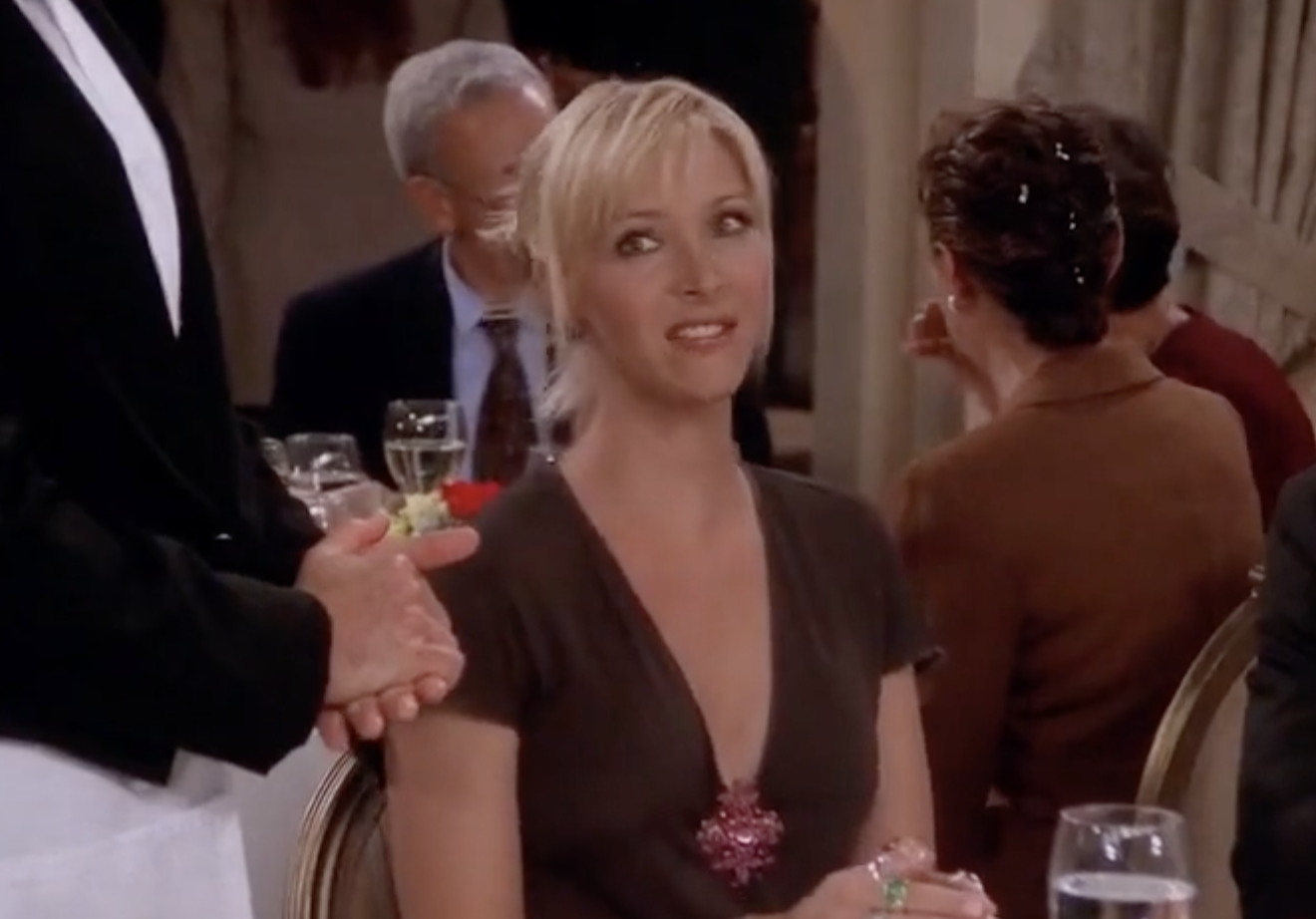}
\caption{Phoebe having a grumpy facial expression while delivering a laughter provoking line.}
\label{phoebe}
\end{figure}

In this case Phoebe's line was delivered with a mean and fed up tone, which was helpful for the model in determining humor. Of course, such a tone is not related to humor in every day speech, so this is an indication of a potential bias in the TV show where such a tone is probably used more often to deliver a punchline of a joke than to actually upset the interlocutor. Even though the multimodal model predicted it correctly, including video modality could strengthen the signal to the model because Phoebe had an uncharacteristically nasty facial expression when uttering her line as seen in Figure \ref{phoebe}.

The both of the models also produce false positives. The following is an example of a dialog that resulted in being falsely labeled as humorous:

\begin{itemize}
\setlength\itemsep{0em}
  \item \textbf{Monica:} Chandler, don't joke with me. Okay? I'm very, very upset right now.
  \item \textbf{Chandler:} Is this the most upset you could be?
  \item \textbf{Monica:} I think so.
\end{itemize}

When this dialog is presented without a wider context, it becomes difficult even for a person to know whether it is supposed to be a joke or not. Monica might very well be talking sarcastically, which is a typical type of humor in the corpus, but in this particular case, she is being sincerely upset. Giving the model more context could alleviate this issue, but of course more context might result in more noise because not all contextual information is relevant. The following is another example where both of the models predicted a false positive:

\begin{itemize}
  \item \textbf{Casting agent:} In your love scene with Sarah... she talks about how she's never seen a naked man who wasn't Jewish.
\end{itemize}

The audience did not laugh after this statement, although in the right context, it might be funny. Here the context was serious. The audience only laughed later on when it became evident that Joey, who was being cast to the movie, did not understand what the agent meant by this utterance. It is clear that what is humorous and not is not always that clear cut especially in a narrow context.

\section{Discussion}

In this section, we discuss our work and the results obtained. As mentioned earlier, one crucial aspect for understanding humor is the context. We have defined the context in our work as a fixed 10 seconds but, in reality, the context might be wider than that. As another attempt to define context, we have split the TV show based on changes in the visual scene. This approach however was not very practical for our needs as it would mistakenly cut scenes based on changes in the camera angel. A potential solution to overcome the problem of fixed contexts is to observe semantic changes for discovering consecutive scenes that share the same topic. We keep this for future work.

Sitcom TV shows are a great resource for computational humor as there is a multitude of humor forms that they present. For this reason, it is challenging for a neural model to capture all humor forms (e.g., irony, sarcasm, satire, exaggeration, personification, silliness, pun and parody), whether they are expressed verbally or visually. Our multimodal model develops its own understanding of what is funny based on the textual and audial features embedded by BERT and HuBERT. Thus, it is incapable of explaining the humorousness it perceives. In other words, the model can say whether something is funny or not, but it cannot say \textit{why} something is funny. A future direction would be to break down these types of humor and feed them collectively to the model, which would enable it to recognize humor from different aesthetics and pinpoint the humor type.

From the error analysis, it is evident that including video modality can help the model in understanding humor by situating what is said to the context presented in the video. However, it is not that clear to know which features would be needed. In Figure \ref{friends_joey-c}, we could see that it is the action that makes the humor more interpretable, whereas in Figure \ref{phoebe} we saw that the facial expression was revealing of humor. If the humor occurs only in the visual modality as seen early on in Figure \ref{friends_joey}, it is in the silliness of how Joey looks while wearing all those clothes. Needless to say, a simple automatically extracted vector representation of the video such as video2vec \cite{hu2016video2vec} is not capable of capturing all these different nuances expressed in the video. It might very well be that an entire TV show does not have enough data for the model to learn to use the video in a meaningful way.

\begin{figure}%
    \centering
    \subfloat[\centering a man and a woman sitting on a bench]{{\includegraphics[width=0.2\textwidth,height=0.125\textwidth]{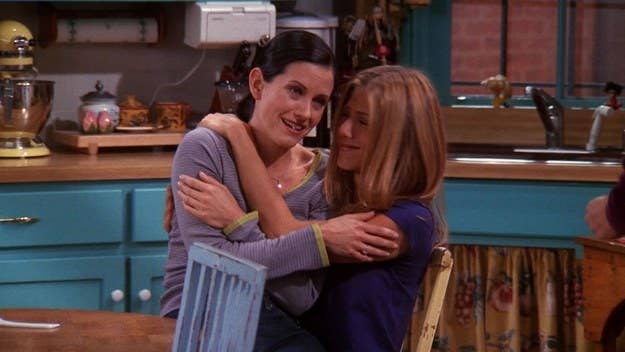} }}%
    \qquad
    \subfloat[\centering a man and a woman sitting at a table]{{\includegraphics[width=0.2\textwidth,height=0.125\textwidth]{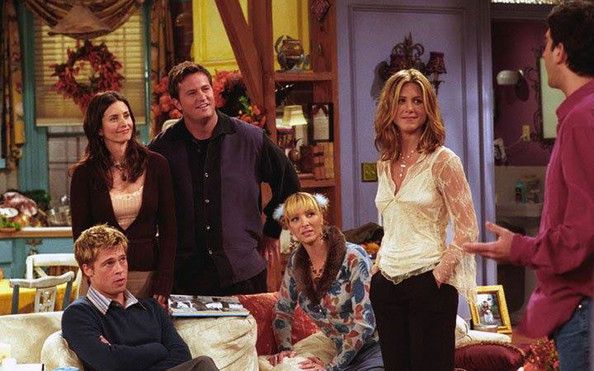} }}%
    \\
    \subfloat[\centering a man and a woman is sitting on a couch with a dog]{{\includegraphics[width=0.2\textwidth,height=0.125\textwidth]{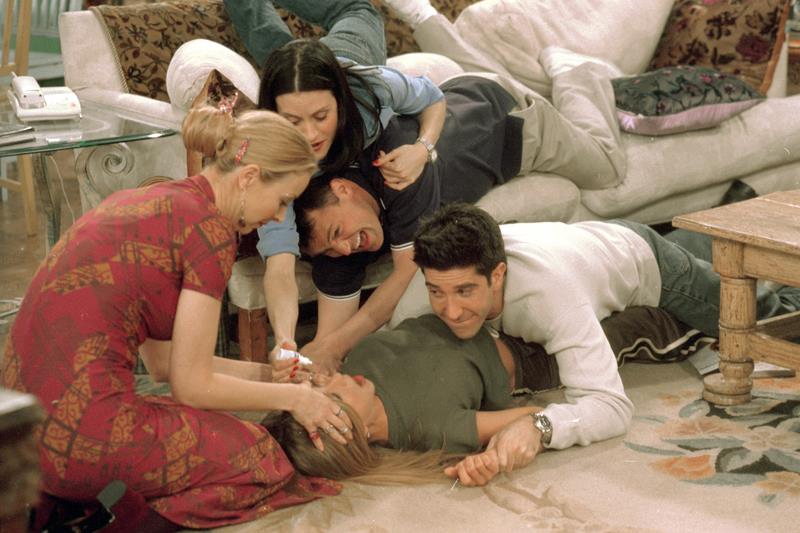} }}%
    \qquad
    \subfloat[\centering a man and a woman playing a video game]{{\includegraphics[width=0.2\textwidth,height=0.125\textwidth]{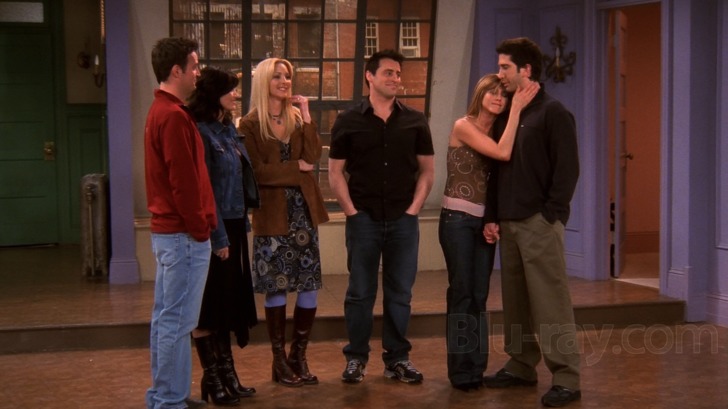} }}%
    \caption{Descriptions produced by the image captioning model}%
    \label{fig:example}%
\end{figure}

We did try to include video features in our models by obtaining textual descriptions of what is happening in a scene by using an existing state-of-the-art video captioning model~\cite{luo2020better}. The idea was that a captioning model could extract relevant information from the video into a textual format that could then be understood by a language model such as BERT. In practice, this turned out to be an impossible task with the current models and the image datasets they were trained on. Figure \ref{fig:example} shows the poor performance the model had on images from \textit{Friends}. In our experiments, we did not see a single correctly captioned image from the TV show while test data from the dataset the model was trained on produced decent results.

We have also seen that the 10 second contextual window is not always enough to resolve whether something is humorous or not. The issue of increasing the context is that more context will also increase the amount of irrelevant context. We do not believe that simply increasing the context from 10 to 20 seconds, for example, is the most optimal way to go about it because some jokes require more context whereas others do not. Perhaps the best way would be to introduce a third model to the pipeline that is trained to determine how much context is needed by identifying how far back in the dialog we can go and still stay in the same topic. A change in topic would indicate that the context goes too far away from what is needed to interpret the joke.

\section{Conclusions}

This work has shown the first steps towards humor interpretation in a multimodal data. Unlike the existing methods, our method does not rely on implicitly marked setup and punchline but can rather detect humor even in cases where the setup of the joke was not made explicit in the text. We have also trained a model that can rank the intensity of humor based on how long the audience laughed. The results are promising and our current research has a lot of potential for future research especially in studying how to deal with video and how much context one should include.

The trained models can be incorporated in other computational creativity models for generating humor. For instance, a system for generating humorous transcripts could utilize our models for determining whether the plot is funny and which version of it would make the audience laugh the most. 

An interesting application of our approach would be to pipeline it with a laughter generator. The models presented in this paper could be used to identify where laughter should be inserted and with what intensity in a comedy show that does not have prerecorded laughter. This could save time in post-editing if used in a professional setting.

Because \textit{Friends} has been translated into multiple languages, this makes it possible to rerun our experiments in different languages with a minimal effort. It also creates an ultimate test-bed for multilingual models where we can test whether a model learning humor from the data in all languages can learn a better representation. 

\bibliography{iccc}

\section*{Acknowledgments}
This work was partially financed by the Society of Swedish Literature in Finland with funding from Enhancing Conversational AI with Computational Creativity, and by the Ella and Georg Ehrnrooth Foundation for Modelling Conversational Artificial Intelligence with Intent and Creativity.

\end{document}